%% file: root.tex
\documentclass[letterpaper, 10 pt, conference]{ieeeconf}  %

\IEEEoverridecommandlockouts                              %

\usepackage{amsmath} %
\usepackage{amssymb}  %
\usepackage{subfiles}
\usepackage{graphicx}
\usepackage{booktabs}       %
\usepackage{multirow}
\usepackage{subcaption}
\usepackage{cite}
\usepackage[export]{adjustbox}
\makeatletter
\let\NAT@parse\undefined
\makeatother
\usepackage[hidelinks,colorlinks=true,linkcolor=black,citecolor=black,urlcolor=blue]{hyperref}

\title{\LARGE \bf
Hearing Touch: Audio-Visual Pretraining \\for Contact-Rich Manipulation
}

\author{Jared Mejia$^{1}$, Victoria Dean$^{2}$, Tess Hellebrekers$^{3}$, Abhinav Gupta$^{1}$%
\thanks{$^{1}$Robotics Institute, Carnegie Mellon University }%
\thanks{$^{2}$Olin College of Engineering}%
\thanks{$^{3}$Meta AI}%
}

\begin{document}

\maketitle
\thispagestyle{empty}
\pagestyle{empty}

\begin{abstract}

\subfile{source/abstract}

\end{abstract}

\section{INTRODUCTION}

\subfile{source/intro}

\section{Related Work}

\subfile{source/related-work}

\section{Manipulation with Audio-Visual Pretraining}
\subfile{source/method}

\section{Experiments}
\subfile{source/experiments}

\section{Conclusion}
\subfile{source/conclusion}

\section{Limitations}
\subfile{source/limitations}

\section*{Appendix}\label{sec:appendix}
\subfile{source/appendix}

\section*{ACKNOWLEDGMENT}

We thank Krishna Suresh, Raunaq Bhirangi, and Mohan Kumar for valuable discussion and feedback, as well as Shaden Naif Alshammari for early work with the contact microphones and Pedro Morgado for advice on using AVID. VD was supported by NSF GRFP and Siebel Scholars. We gratefully acknowledge the support of ONR MURI.

\bibliographystyle{IEEEtran}

\end{document}

%% file: source/abstract.tex
Although pre-training on a large amount of data is beneficial for robot learning, current paradigms only perform large-scale pretraining for visual representations, whereas representations for other modalities are trained from scratch. In contrast to the abundance of visual data, it is unclear what relevant internet-scale data may be used for pretraining other modalities such as tactile sensing. Such pretraining becomes increasingly crucial in the low-data regimes common in robotics applications. In this paper, we address this gap by using contact microphones as an alternative tactile sensor. Our key insight is that contact microphones capture inherently audio-based information, allowing us to leverage large-scale audio-visual pretraining to obtain representations that boost the performance of robotic manipulation. To the best of our knowledge, our method is the first approach leveraging \emph{large-scale multisensory pre-training} for robotic manipulation. For supplementary information including videos of real robot experiments, please see \href{https://sites.google.com/view/hearing-touch}{https://sites.google.com/view/hearing-touch}.

%% file: source/intro.tex
Two key components consistently improve the performance of robotic manipulation: (1) pre-training on a large amount of data \cite{nair2022r3m, ma2022vip, majumdar2023we, ebert2021bridge, kumar2022pre} and (2) using multisensory input, especially tactile sensing \cite{li2022see, zhang2019leveraging, calandra2018more, calandra2017feeling, lee2019making, murali2018learning}. While recent work has leveraged pretraining on large-scale video datasets to create reusable \textit{vision} representations for robot learning \cite{nair2022r3m, ma2022vip, majumdar2023we}, there has been little focus on large-scale pretraining for other modalities such as tactile sensing. This gap arises due to the lack of relevant data at a comparable scale for tactile sensing. As a result, current approaches using non-visual sensory modalities are restricted to learning from a limited amount of task-specific data \cite{lee2019making, thankaraj2022sounds}. How can we leverage internet data in pretraining tactile representations for manipulation?  

Piezo contact microphones have emerged as a promising sensor in robotics due to their ability to capture high-frequency temporal information through structural vibrations captured as audio. Prior work has already demonstrated the ability to use contact audio for manipulation tasks \cite{li2022see, thankaraj2022sounds, clarke2018learning}. In contrast to traditional tactile sensors, the signal provided by contact microphones is inherently audio; hence recent work on learning audio-visual representations may apply to contact audio obtained from robot interactions.

\begin{figure}
\centering
\includegraphics[trim=0 0 0 0, clip, width=\linewidth]{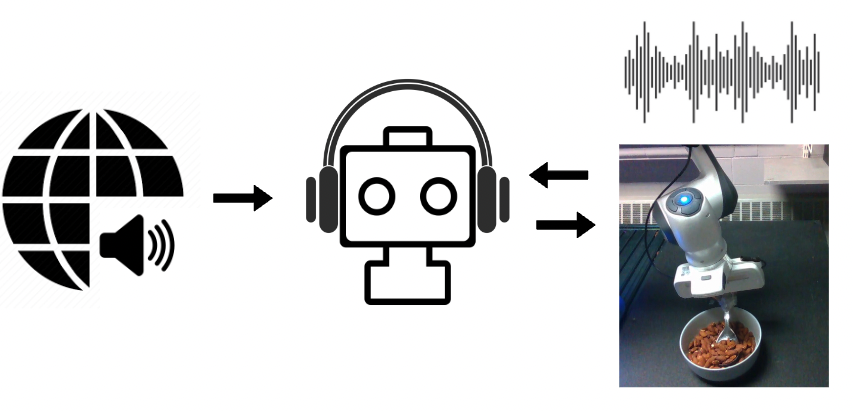}
\caption{\textbf{Hearing touch}: We enable multisensory pretraining for manipulation by transferring audio-visual representations to manipulation tasks using vision and contact audio.}
\label{fig:teaser}
\end{figure}

We investigate how large-scale audio-visual training might be beneficial for learning contact audio representations for robotic manipulation. Our method makes use of Audio-Visual Instance Discrimination (AVID) \cite{morgado2021audio}, a self-supervised learning approach to learn audio-visual representations, pre-trained on Audioset \cite{gemmeke2017audio}, a dataset containing over 2 million 10-second video clips of human and animal sounds, music, and environmental sounds drawn from the internet. Initializing our encoder with AVID weights, we train a policy with behavior cloning that fuses visual and audio inputs with self-attention in order to predict actions.

We validate our approach with experiments on three real-world manipulation tasks in the low-data regime, using at most 60 demonstrations per task. Surprisingly, despite the domain gap between the audio in Audioset and contact audio obtained through manipulation, we find that our approach improves performance over visual-only policies---especially in test settings where objects and locations differ significantly from the training data. Furthermore, our approach outperforms equivalent policies with audio encoders trained from scratch. Our experimental results reveal a promising avenue for multimodal pretraining across many robotic applications where neither vision alone nor training multisensory representations from scratch are sufficient.

%% file: source/related-work.tex
\begin{figure*}[ht]
\vspace{0.5em}
\centering
    \includegraphics[width=.75\textwidth]{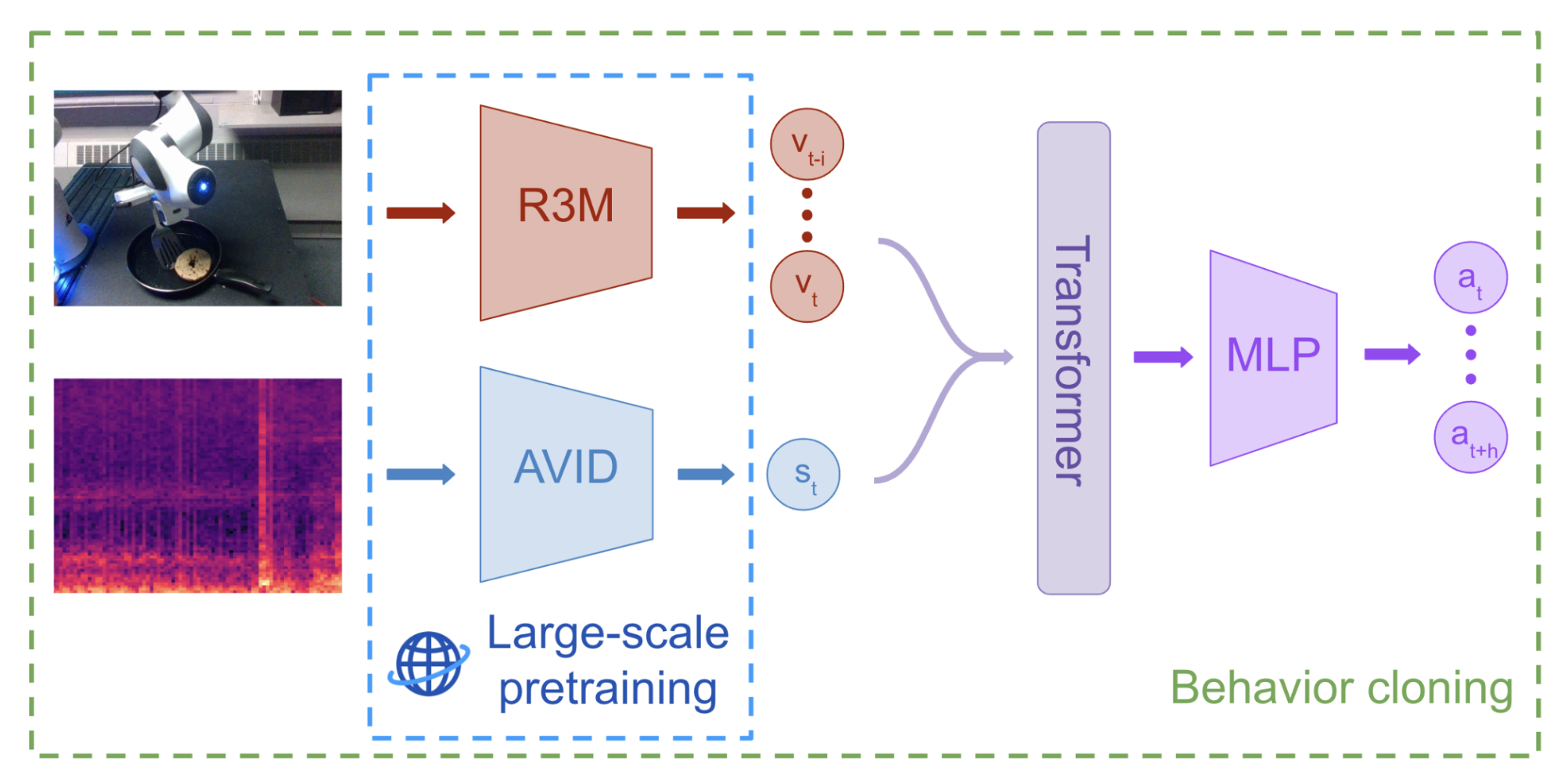}
    \caption{\textbf{Two-stage model training.} AVID and R3M pretraining leverages the large scale of internet video data (blue dashed box). We initialize the vision and audio encoders with the resulting pre-trained representations and then train the entire policy end-to-end with behavior cloning from a small number of in-domain demonstrations. The policy takes image and spectrogram inputs (left) and outputs a sequence of actions in delta end effector space (right).}
\label{fig:method}
\end{figure*}

\paragraph{Audio in robotics}
Several works have shown the ability to reason over audio in robotics scenarios including object recognition \cite{gandhi2020swoosh}, material classification \cite{clarke2022diffimpact}, estimating the volume and flow of granular material \cite{clarke2018learning}, exploration in RL \cite{dean2020see}, occluded manipulation \cite{du2022play}, manipulation for sound replication \cite{thankaraj2022sounds}, and waypoint setting in audio-visual navigation \cite{chen2020learning}. 
SHF \cite{li2022see} introduces a mechanism for fusing input from a camera, a Gelsight sensor \cite{yuan2017gelsight}, and a contact microphone attached to the object of interest with self-attention for manipulation. Though our method also uses self-attention to fuse multisensory representations, we focus on leveraging large-scale audio pretraining, using visual input from a third-person camera and a contact microphone mounted directly on the robot enabling the robot to reason over vibrations caused by contact between tools and objects.

\paragraph{Tactile sensing for manipulation}
Several types of tactile sensors exist for application to robotic manipulation \cite{lambeta2020digit, li2019elastomer, bhirangi2021reskin, donlon2018gelslim, sundaralingam2019robust, bhattacharjee2013tactile}. We use contact microphones as an alternative tactile sensor, which are relatively inexpensive in comparison to common tactile sensors and can record vibrations with up to 1000 times higher frequency than optical and magnetic-based tactile sensors (32-48 kHz vs 30-400 Hz) \cite{lambeta2020digit, li2019elastomer, bhirangi2021reskin}. Recent work has focused on applying traditional tactile sensors for learning to grasp objects without visual observations \cite{murali2018learning} and in combination with visual observations for learning to improve the grasp of an object \cite{calandra2018more}. Our method using contact audio allows the sensor to measure vibrations directly via the sensor mounted on the gripper as well as indirectly through vibrations traveling along tools grasped by the gripper.

\paragraph{Audio-visual representation learning}
Self-supervised representation learning has been applied to the audio-visual domain, using audio-visual correspondence (AVC) as a form of cross-modal self-supervision from video \cite{arandjelovic2017look, arandjelovic2018objects}. Other approaches make use of the synchronization between vision and sound for sound representations \cite{aytar2016soundnet}, audio-visual sound separation \cite{zhao2019sound}, and sound localization \cite{chen2021localizing}. More recent work explores contrastive learning methods to discriminate between training instances using cross-modal and within-modal targets \cite{morgado2021audio, morgado2021robust, patrick2020multi}. In our work, we use a pre-trained implementation of AVID \cite{morgado2021audio} for obtaining audio-visual representations.

\paragraph{Representation learning for robotic manipulation}
Several recent works use self-supervision to decouple representation learning of sensory inputs from behavior learning for robotic manipulation tasks \cite{radosavovic2023real,  thankaraj2022sounds, pari2021surprising, lee2019making, arunachalam2022dexterous}. A recent trend aims to obtain a universal visual representation---a single perception module pre-trained on large amounts of video data that can be frozen and used for downstream policy learning \cite{nair2022r3m, ma2022vip, majumdar2023we}, however, there has been little focus on large scale pre-training for representation learning beyond vision in the context of robot manipulation.
That Sounds Right \cite{thankaraj2022sounds} also explores contact audio pre-training for behavior learning, however, their approach utilizes self-supervised learning using only task-specific data, whereas our method leverages the richness and diversity of large-scale audio-visual data for pre-training a contact audio representation. Further, we operate in the low-data regime with less than 100 demonstrations per task, whereas \cite{thankaraj2022sounds} collects 5,000 data points per task. We demonstrate the benefit of large-scale pre-training over in-domain SSL in the low-data setting.

%% file: source/method.tex
Given the difficulty and expense of collecting data in robotic settings, we turn toward leveraging more easily attainable large-scale sources of information such as internet data for learning manipulation policies. By utilizing contact microphones, we move beyond pre-training solely for visual input and obtain a means of pre-training a tactile sensor with large amounts of rich, audio-visual data. We outline further details of our approach in the following sections.

\subsection{Sensors}
\begin{figure*}[t]
\vspace{1em}
\centering
    \begin{subfigure}[b]{.20\textwidth}
        \centering
        \adjincludegraphics[height=3.9cm,trim={0 0 {.75\width} 0},clip]{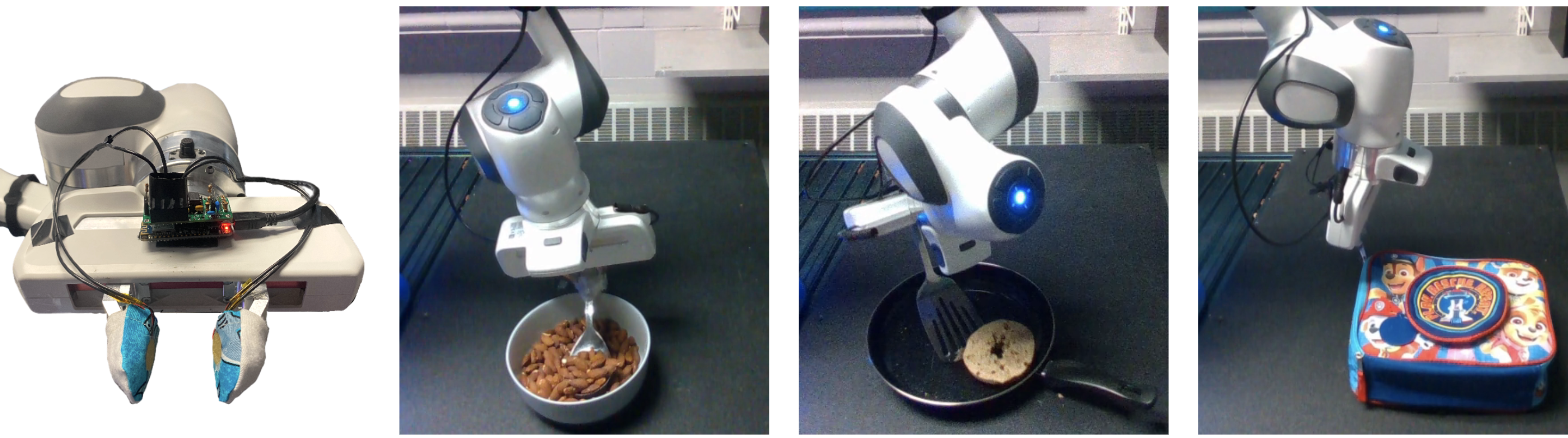}
        \caption{Hardware Setup}
        \label{fig:hardware}
    \end{subfigure}
    \begin{subfigure}[b]{0.20\textwidth}
        \centering
        \adjincludegraphics[height=3.9cm,trim={0 0 0 0},clip]{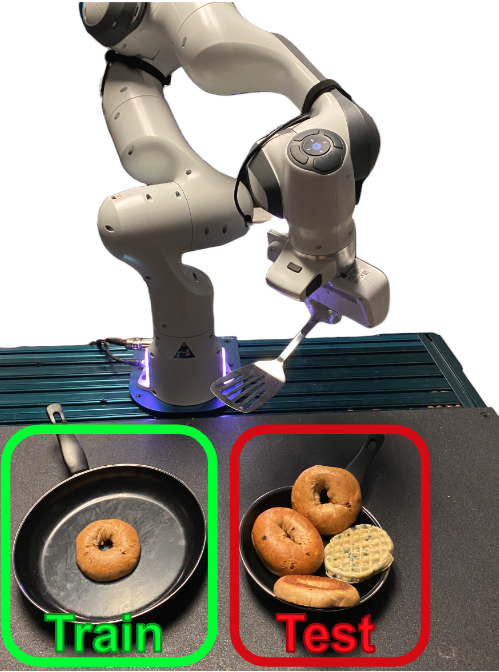}
        \caption{Flipping task}
        \label{fig:flipping-train-test}
    \end{subfigure}
    \begin{subfigure}[b]{0.25\textwidth}
        \centering
        \adjincludegraphics[height=3.9cm,trim={0 0 0 0},clip]{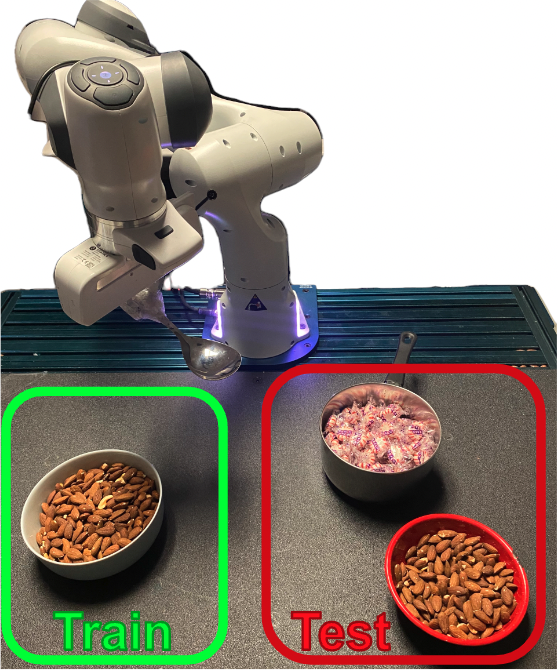}
        \caption{Scooping task}
        \label{fig:scooping-train-test}
    \end{subfigure}
    \begin{subfigure}[b]{0.28\textwidth}
        \centering
        \adjincludegraphics[height=3.9cm,trim={0 0 0 0},clip]{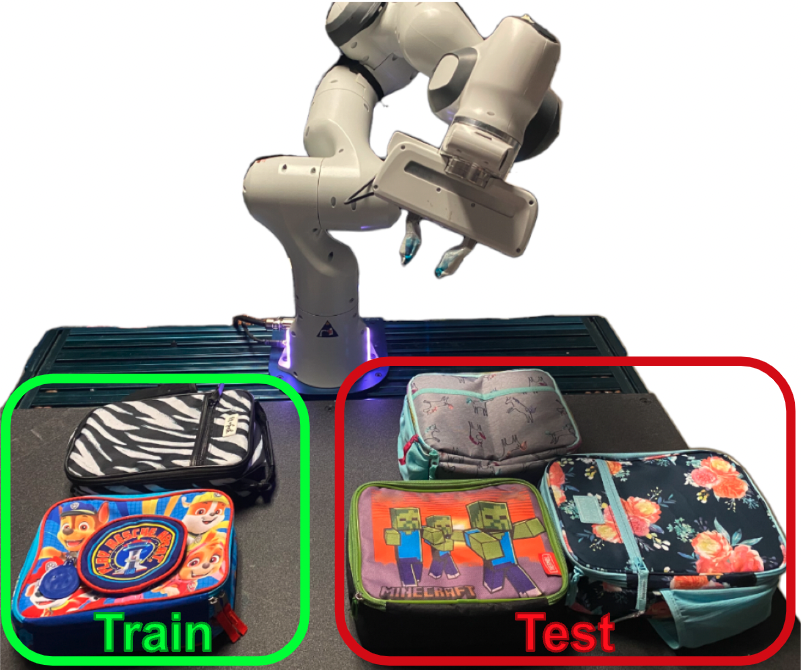}
        \caption{Zipping task}
        \label{fig:zipping-train-test}
    \end{subfigure}
    
    \caption{\textbf{Hardware and task setup.} We attach the Piezo contact microphones to our gripper to record vibrations in the form of audio and run experiments on three real-world tasks with significant visual differences between train and test settings.} \label{fig:tasks}
\end{figure*}

At each timestep, we collect image observations $v_t$ and two-second clips of contact audio $a_t$. Image observations are obtained from a third-person view camera and audio is obtained by averaging the signal captured from four contact microphones mounted on the robot. Contact microphones capture vibrations, they are sensitive to contact not only directly between objects and the sensors but also contact resulting in vibrations traveling between objects. As a result, our setup allows the robot to sense subtle interactions between surfaces and tools that are grasped by the arm, as in the flipping task which requires the use of a spatula, and the scooping task requiring the use of a spoon (Section \ref{sec:setup}).

\subsection{Audio and Visual Representation Pretraining}
Our method uses large-scale audio-visual pre-training to initialize our audio encoder and large-scale visual pre-training to initialize our visual encoder. The audio encoder is extracted from AVID \cite{morgado2021audio} pre-trained on audio-visual pairs from Audioset \cite{gemmeke2017audio} with cross-modal discrimination, encouraging the network to learn video features that match the corresponding audio features and vice-versa. To isolate the effect of large-scale pre-training for our audio encoder, we use R3M \cite{nair2022r3m}, a proven method for pre-training visual features in robotic applications, R3M, with a ResNet18 \cite{he2016deep} pre-trained on Ego4D human video dataset \cite{grauman2022ego4d} with time contrastive learning and video-language alignment.
Following \cite{dean2022dont}, we keep both encoders unfrozen, continuing to update the weights during policy learning. 

\subsection{Audio-Visual Behavior Cloning}
We train a policy with behavior cloning on a small number of in-domain demonstrations (described in Section \ref{sec:setup}). The model architecture is visualized in Fig. \ref{fig:method}. At each timestep, the policy takes in a two-second audio clip $s_t$ and a sequence of $i$ images $v_{t-i}, \dots, v_{t}$ spanning the same two-second window, which are fed through the audio and image encoders, respectively. We apply learned positional embeddings to each of the encoded representations and pass the result as input to a transformer decoder network similar to \cite{li2022see}. Similar to \cite{zhou2023train, chi2023diffusion} our method is quasi open-loop---at time step $t$ the policy predicts $H$ steps of actions, of which $h \leq H$ steps of actions are executed on the robot without re-planning. This approach allows the policy to remain responsive to subtle changes in the audio input while encouraging temporal action consistency and mitigating the effect of non-Markovian behaviors such as pauses in demonstrations. In particular, the final component of our network is a multi-layer perceptron that outputs actions $a_t, \dots, a_{t+h}$ over a short horizon of $h$ timesteps. Here, each action $a_t$ is a 6-dimensional continuous delta-end effector action composed of the Cartesian displacement $(x, y, z)$ and the change in Euler angles $(\alpha, \beta, \gamma)$.  We optimize the network to minimize the standard MSE loss $\ell = \frac{1}{H} \sum_{j=0}^{H} (a_{t+j} - \pi(v_{t-i}, \dots, v_{t},  s_t)_j)^2$. Please see Section \ref{sec:appendix} for more architectural details.

%% file: source/experiments.tex
In our experiments, we aim to answer two key questions: (1) Do contact microphones mounted on a robot arm capture interactions difficult to perceive with vision alone? (2) Does large-scale pre-training for audio-based tactile sensors yield representations that are useful for robot manipulation?

\begin{figure*}[t]
\vspace{1em}
\centering
\adjincludegraphics[height=4.9cm,trim={0 0 0 0},clip]{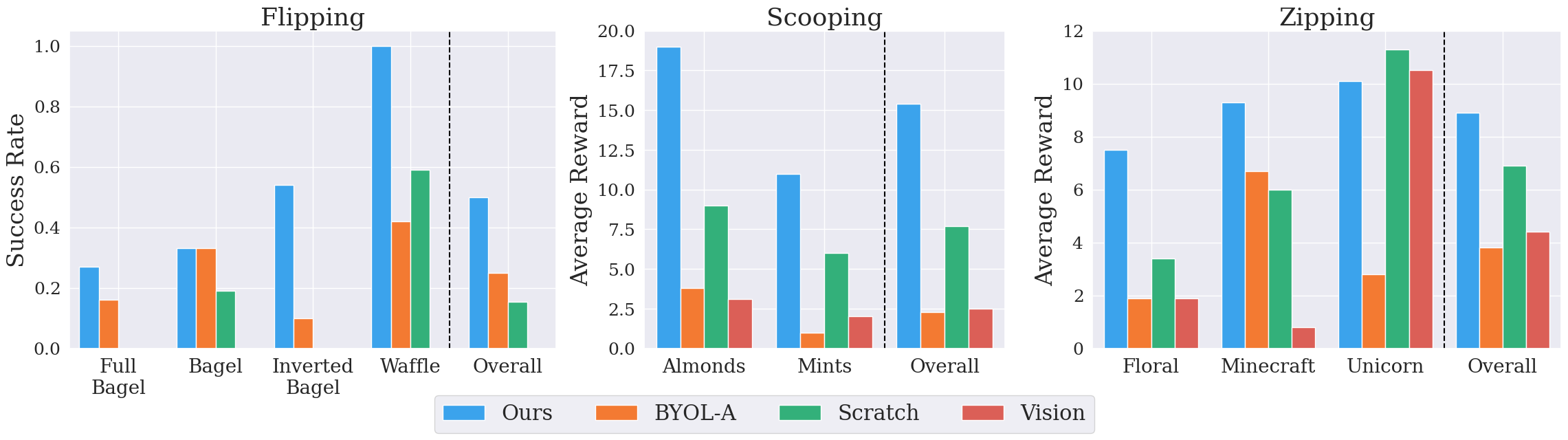}
\caption{\textbf{Success rates across methods and tasks.} Our method, shown in blue, outperforms baselines in all but one setup of the zipping task. Furthermore, our method displays much less variation in performance between different configurations of each task, showcasing an increase in the ability to generalize to drastic visual differences as a result of learning useful audio representations.}
\label{fig:plots}
\end{figure*}

We address these questions through real-world experiments on our setup described in Section \ref{sec:setup} by evaluating across three tasks (Section \ref{sec:tasks}) and four methods (Section \ref{sec:baselines}) in the low-data setting under conditions requiring significant generalization beyond the training data.

\subsection{Setup} \label{sec:setup}
\paragraph{Hardware}
We control a Franka Emika Panda Arm using an inverse kinematics solver to convert 6-DoF delta end effector Cartesian position and Euler rotation input to 7-DoF joint action. The end effector actions are commanded at 30 Hz. On the Franka gripper, we mount four Piezo contact microphones, each of which records audio at 32 kHz. We use an Intel D435 RealSense camera with a fixed third-person view to collect image observations at 30 Hz.

\paragraph{Data Collection}
Demonstrations are collected via teleoperation using an Oculus Quest headset. The visual data collected by the Intel D435 RealSense camera collects images with a resolution of $480 \times 640$. The audio waveforms are averaged across the four sensors and downsampled to 16 kHz. We normalize the audio waveforms and generate mel spectrograms of the 2s audio segment following the audio preprocessing in \cite{morgado2021audio}.

\subsection{Tasks}
\label{sec:tasks}
We present experiments on three real-world manipulation tasks, shown in  Fig. \ref{fig:hardware}. The zipping task demonstrates the contact microphone's abilities to directly record vibrations touching the gripper, while the flipping and scooping tasks show their ability to record indirect contacts through vibrations traveling along tools (the spoon and spatula). We train on 40, 60, and 50 demonstrations for the flipping, scooping, and zipping tasks, respectively. %

\subsection{Baselines and Implementation Details}
\label{sec:baselines}
We conduct experiments with our method and three other baselines. We use different methods of pretraining in order to measure the effect of large-scale audio-visual pretraining on learning a useful contact audio representation for manipulation. All methods incorporating audio use the same architecture: R3M \cite{nair2022r3m} pre-trained on Ego4d \cite{grauman2022ego4d} with a ResNet18 \cite{he2016deep} backbone to initialize the image encoder. 
\begin{itemize}
    \item \textbf{Vision-Only:} a baseline that shares the same architecture as our method, except that it only uses image frames as input. This baseline tests whether the signal from contact microphones is beneficial in our setup.
    \item \textbf{Scratch:} a baseline with randomly initialized weights for the audio encoder. This baseline tests how contact audio pretraining affects performance.
    \item \textbf{BYOL-A:} Bootstrap Your Own Latent for Audio (BYOL-A) \cite{niizumi2021byol}, a self-supervised approach to learning audio representations using only in-domain data.
    This baseline compares the effect of large-scale audio-visual pre-training to in-domain audio pre-training, with an emphasis on the \emph{amount} of pre-training data.
\end{itemize}

\subsection{Results}

\setlength\tabcolsep{3.5pt} %

\begin{table}[b]
\caption{Rewards and success rates across tasks.}
\vspace{-1em}
\label{tab:results}
\begin{center}
\begin{tabular}{lcccccc}
\toprule
  & \multicolumn{1}{c}{Flipping} & \multicolumn{2}{c}{Scooping} & \multicolumn{2}{c}{Zipping} \\
  \cmidrule(lr){2-2} \cmidrule(lr){3-4} \cmidrule(lr){5-6}
& \rule{0pt}{2ex} Success \% & Reward & Success \% & Reward & Success \%\\
\midrule
\texttt{Ours}       & \textbf{50.0\%} & \textbf{15.4} & \textbf{78.1\%} & \textbf{8.9} & \textbf{88.9\%} \\
\texttt{BYOL-A}      & 25.0\% & 2.3 & 25.0\% & 3.8 & 66.7\% \\
\texttt{Scratch}     &  15.4\% & 7.7 & 50.0\% & 6.9 & 72.2\% \\
\texttt{Vision-Only}    &  0.0\% & 2.5 & 28.1\% & 4.4 & 44.4\% \\
\bottomrule
\end{tabular}
\end{center}
\end{table}

\begin{figure*}[t]
    \vspace{1em}
    \centering
    \begin{subfigure}[b]{0.24\textwidth}
        \centering
        \adjincludegraphics[height=3cm,trim={0 0 0 0},clip]{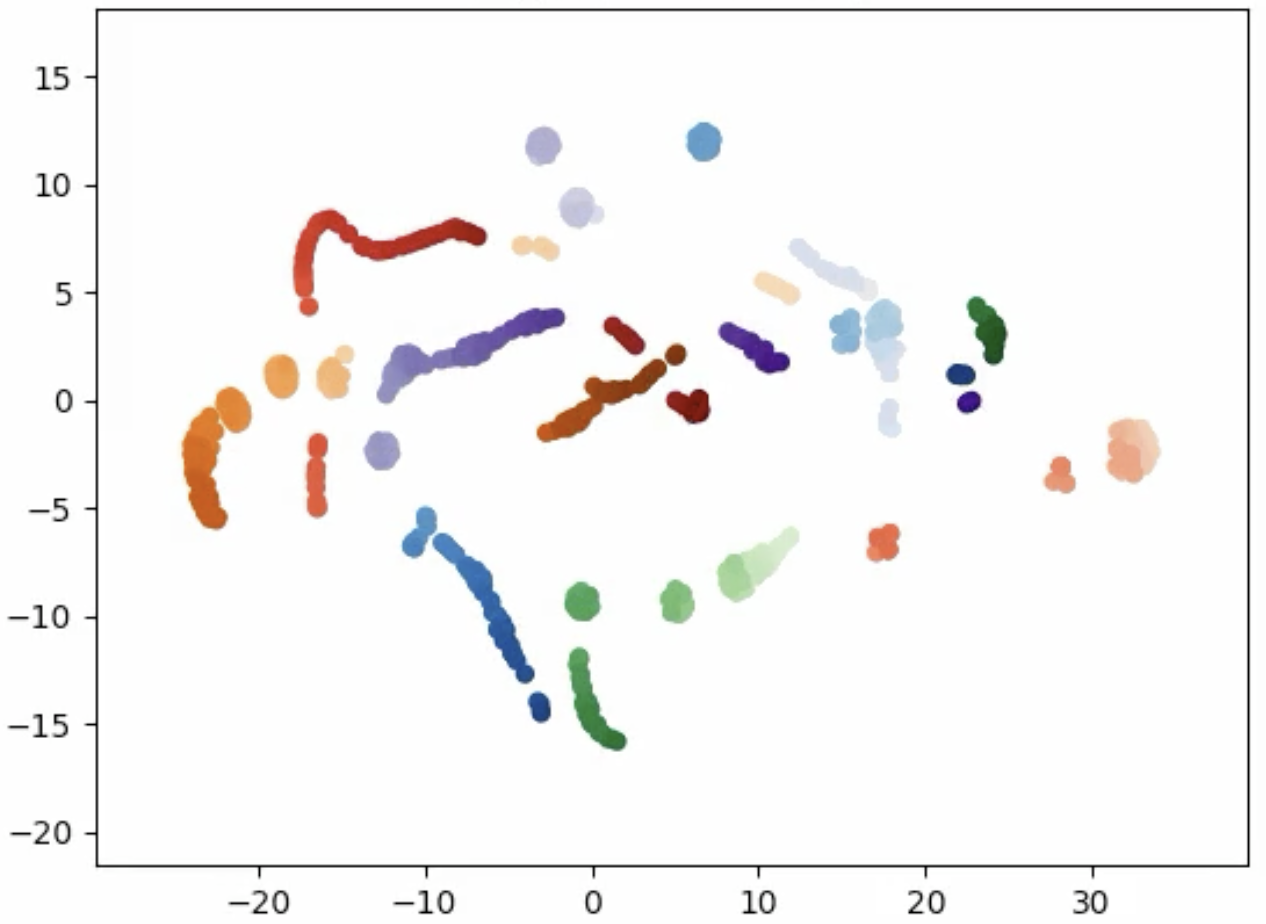}
        \caption{Vision-Only}
        \label{fig:t-sne-vision-only}
    \end{subfigure}
    \begin{subfigure}[b]{0.24\textwidth}
        \centering
        \adjincludegraphics[height=3cm,trim={0 0 0 0},clip]{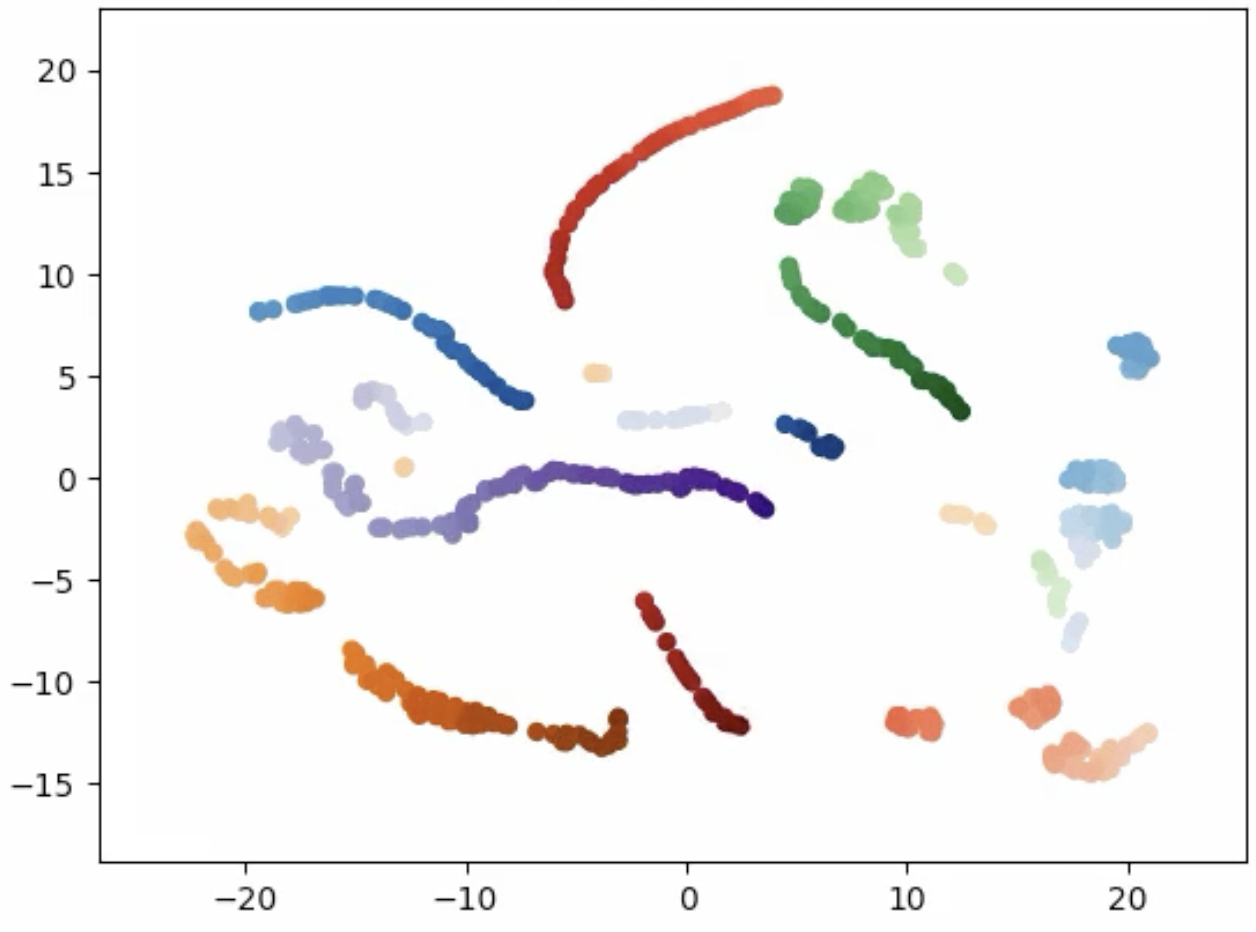}
        \caption{Scratch}
        \label{fig:t-sne-scratch}
    \end{subfigure}
    \begin{subfigure}[b]{0.24\textwidth}
        \centering
        \adjincludegraphics[height=3cm,trim={0 0 0 0},clip]{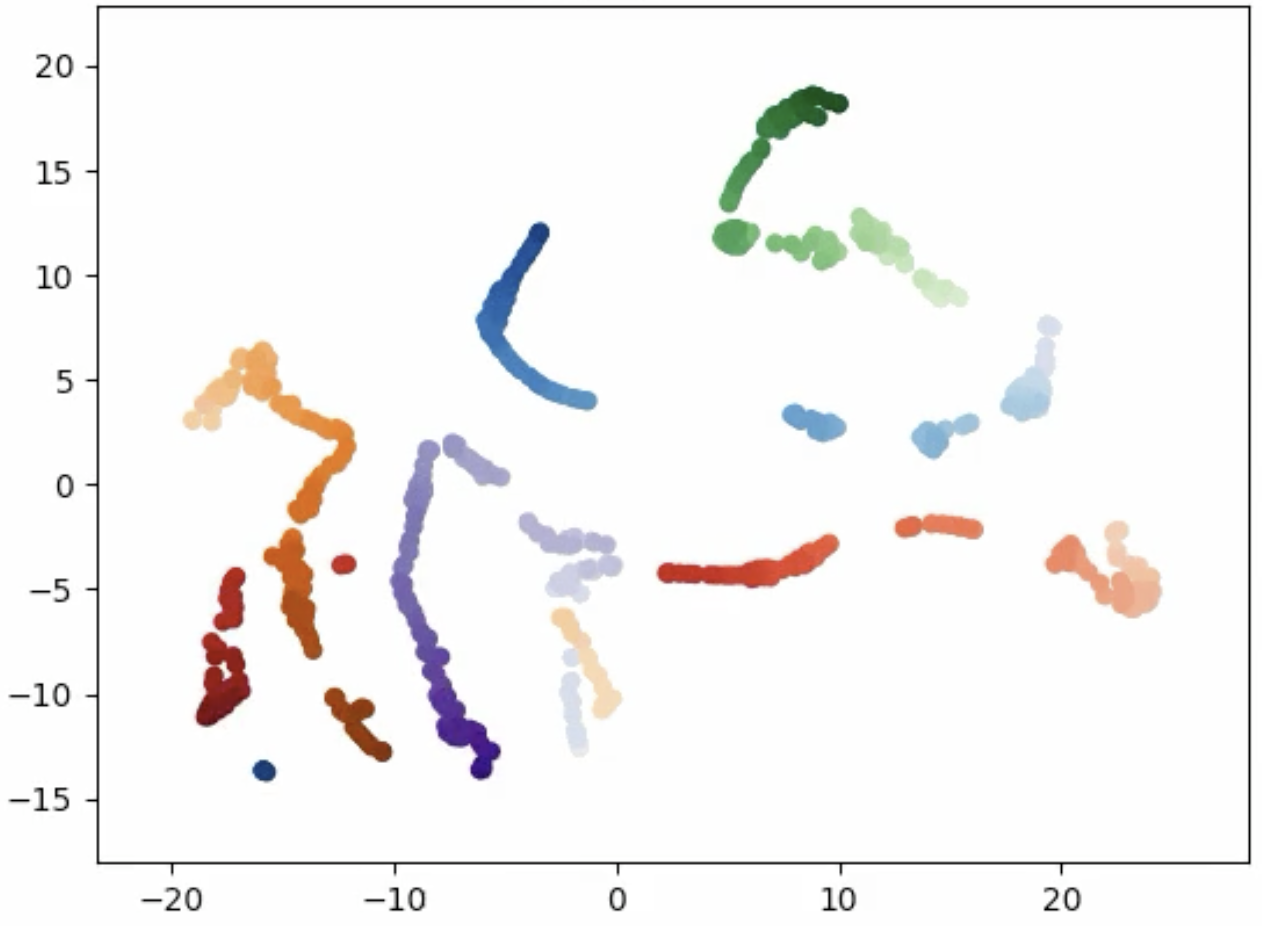}
        \caption{BYOL-A}
        \label{fig:t-sne-byol-a}
    \end{subfigure}
    \begin{subfigure}[b]{0.24\textwidth}
        \centering
        \adjincludegraphics[height=3cm,trim={0 0 0 0},clip]{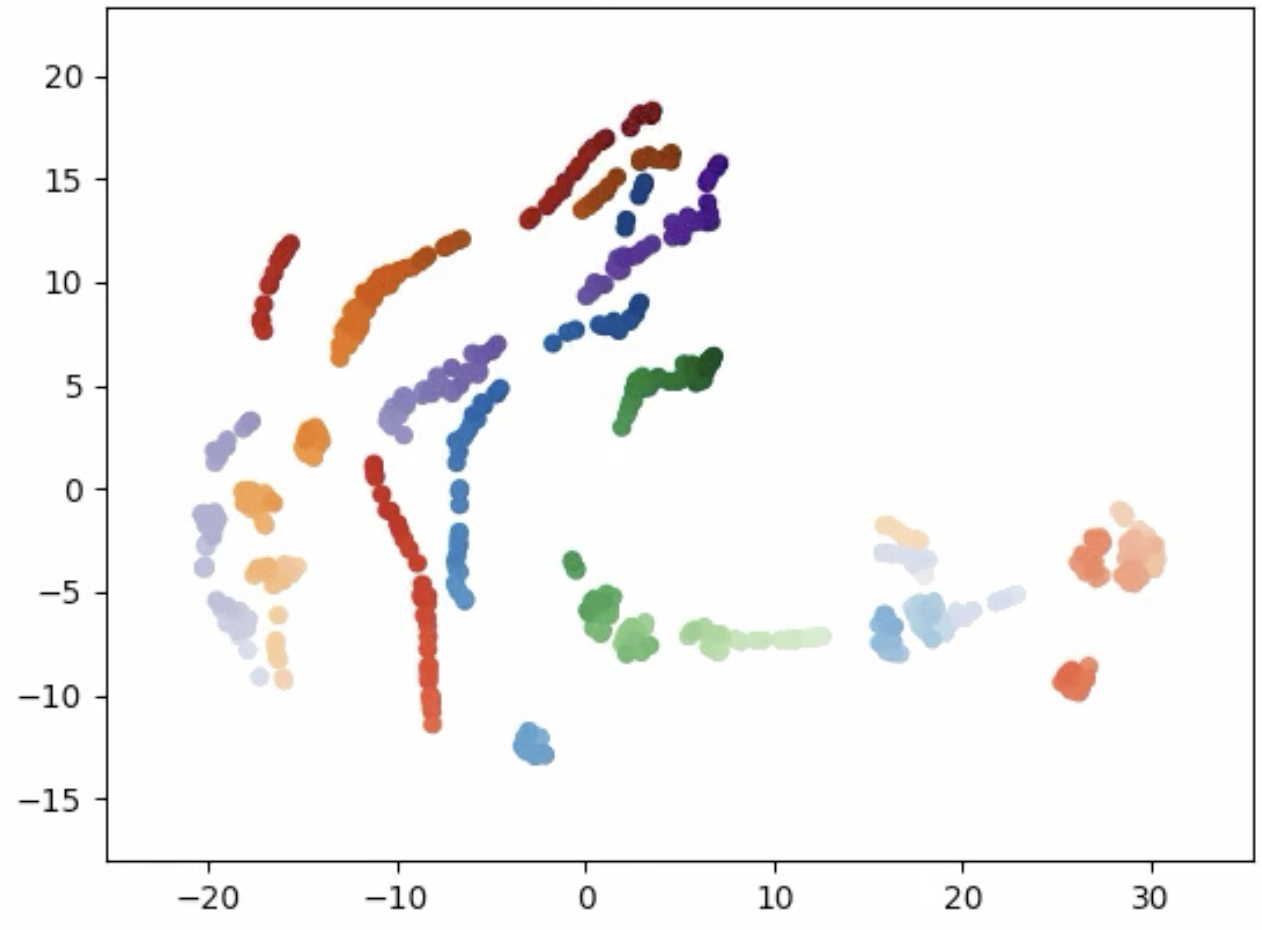}
        \caption{Ours}
        \label{fig:t-sne-ours}
    \end{subfigure}
    \begin{subfigure}[b]{0.72\textwidth}
        \centering
        \includegraphics[width=\textwidth]{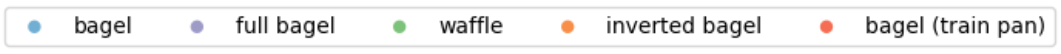}
    \end{subfigure}
    
    \caption{\textbf{t-SNE 2D projection.} For comparative analysis of the learned embedding spaces, we visualize projections of the learned representations from each method in each variation of the flipping task. Lighter hues indicate the starting points and darker hues indicate the end points of the trajectories. Please see the video on our \href{{https://sites.google.com/view/hearing-touch}}{website} for a better visualization.}
    \label{fig:t-sne}
\end{figure*}

The evaluation results across different variations the tasks are visualized in Fig. \ref{fig:plots} and summarized in Table \ref{tab:results}. Our method using large-scale audio-visual pre-training outperforms all baselines across each of the three tasks with an average $23\%$ higher 0-1 success rate and an average $76\%$ increase in reward against the next best-performing baseline. Further, our method outperforms or matches the performance of all baselines in $8/9$ tasks, displaying a lower variation in performance between different configurations of each task, indicating greater robustness to visual features.

The Vision-Only baseline yields the worst performance across all tasks, providing evidence that contact audio improves the performance of manipulation policies over vision alone. Between BYOL-A and Scratch, the results are mixed---in the Flipping task BYOL-A outperforms Scratch and in Scooping and Zipping, Scratch performs better. Although BYOL-A includes an additional pre-training phase, the comparable performance with Scratch suggests that the augmentation techniques used by BYOL-A, while useful for learning audio representations for audio classification tasks when pre-trained on large audio datasets \cite{niizumi2021byol}, are not effective when restricted to a small set of contact audio for learning manipulation policies. In contrast, our method utilizing AVID pre-training on Audioset greatly improves performance over Scratch and BYOL-A, demonstrating that the large-scale aspect of our method's audio-visual pre-training is the component most crucial to its success.

\subsubsection{Qualitative Analysis}

Many of the configurations of the task are difficult due to the noticeable visual differences between the train and test settings. As a result, the baselines suffer heavily from the domain shift and fail to generalize, often moving in jerk motions or away from the object of interest, even before coming into contact with objects. In contrast, our method appears to suffer less from the significant visual differences, suggesting that a good audio representation may prevent the model from overfitting to visual features during training. 

The Vision-Only approach suffers most from the inability to perceive subtle interactions between surfaces, such as whether the spatula has successfully been slid under the bagel or the zipper is stuck on a corner. Despite having access to the same information as our method, the BYOL-A and Scratch baselines fail to reason effectively over the audio and utilize the additional information for taking action. 

In the scooping task, our method consistently learns to push the spoon deeper into the bowl until contact is made with the edge, and then tilt the spoon upward as the edge drags along the side of the bowl, increasing the amount of material scooped. This is more like the behavior of the training data than the baselines, which often fail to begin digging the spoon into the material as a result of misestimating the depth and relying on vision alone or scooping too shallow.

\subsubsection{t-SNE Visualizations}

To better understand the learned representations of our method in comparison with the baselines, we visualize 2D projections of the transformer output embeddings using t-SNE initialized with PCA deterministically. For each method, we plot the projections of the embeddings from a sample trajectory over time for each variation of the flipping task, including both train and test settings (Fig. \ref{fig:t-sne}). For our method, although the representations are spaced apart at the beginning of the trajectories likely due to the visual differences across settings (bottom right corner), the projections converge over the course of trajectories (moving clockwise) as the flipping motion is performed and completed. The visualization suggests the audio representations learned as a result of large-scale pre-training allow for the attention mechanism to better combine the audio-visual tokens, resulting in a more well-structured embedding space in comparison with the baselines. %

\subsection{Ablation Studies}

\subsubsection{Zero-Shot Transfer}\label{sec:zero-shot-transfer}
\begin{figure*}
    \vspace{1em}
    \centering
    \begin{subfigure}[b]{0.24\textwidth}
        \centering
        \adjincludegraphics[height=3.6cm,trim={0 0 0 0},clip]{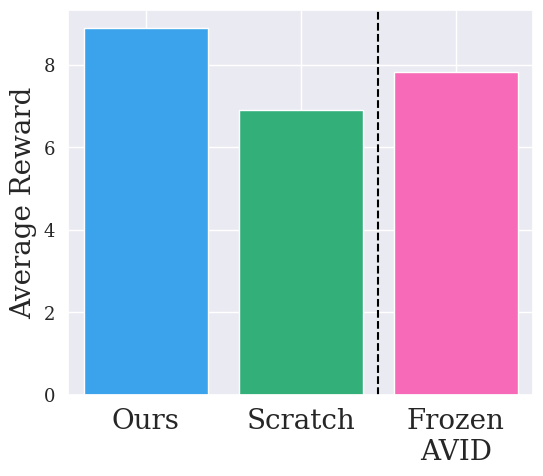}
        \caption{Zero-Shot Transfer}
        \label{fig:zero-shot-transfer}
    \end{subfigure}
    \begin{subfigure}[b]{0.24\textwidth}
        \centering
        \adjincludegraphics[height=3.6cm,trim={0 0 0 0},clip]{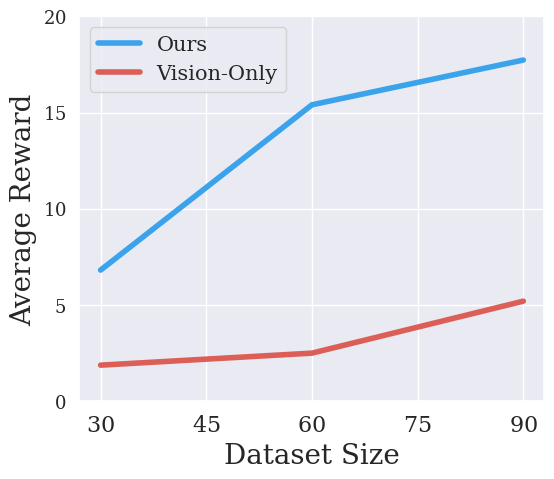}
        \caption{Scaling Performance}
        \label{fig:scaling-performance}
    \end{subfigure}
    \begin{subfigure}[b]{0.24\textwidth}
        \centering
        \adjincludegraphics[height=3.6cm,trim={0 0 0 0},clip]{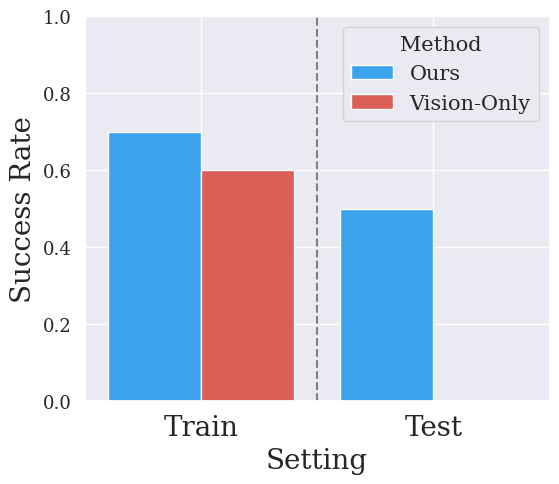}
        \caption{Generalization}
        \label{fig:generalization-vision}
    \end{subfigure}
    \begin{subfigure}[b]{0.24\textwidth}
        \centering
        \adjincludegraphics[height=3.6cm,trim={0 0 0 0},clip]{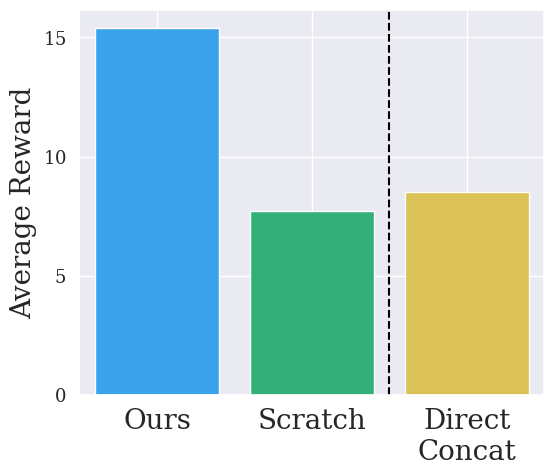}
        \caption{Architecture Ablation}
        \label{fig:arch-ablation}
    \end{subfigure}
    \caption{\textbf{Ablations}. We evaluate the zero-shot transfer of frozen pre-trained audio representations (a), the effect of dataset size (b), the generalization ability of our method (c), and the importance of self-attention to fuse sensory features (d).}
    \label{fig:ablations}
\end{figure*}

To get a better sense of how relevant pre-trained AVID weights are to downstream manipulation tasks, we train a version of our method with frozen AVID weights during policy learning (Fig. \ref{fig:zero-shot-transfer}). The results show that keeping the pre-trained audio encoder weights frozen during policy learning only slightly diminishes the performance of our method and still outperforms the next best baseline on the zipping task, highlighting the applicability of the general sensory representations learned from large-scale internet data for downstream manipulation tasks.

\subsubsection{Scaling Performance}
We run evaluations on the scooping task for models trained with dataset sizes $50\%$ (30 demos) and $150\%$ (90 demos) of the original data after collecting more demonstrations. As shown in Fig. \ref{fig:scaling-performance}, our method continually improves at a steady rate with increasing training data size, roughly matching the rate of improvement for the Vision-Only baseline.

\subsubsection{Generalization}
To further investigate the poor performance of the Vision-Only baseline in comparison to our method on the flipping task, we compare the results between the train and test settings (Fig. \ref{fig:generalization-vision}). The success rate of both methods is closer under the train settings, with our method performing $10\%$ better.
Evidently, despite using image augmentations during training, the Vision-Only baseline overfits to the visual features of the train settings in the demonstration data, resulting in a $60\%$ drop in performance when applied to the test settings. In contrast, our method only sees a drop in success rate of about $20\%$ between train and test settings, suggesting that pre-trained audio features prevent the network from overfitting to visual details in the training setting, hence attaining better generalization abilities. 

\subsubsection{Architecture Ablation}
We replace the transformer with an MLP including an added additional linear layer to ensure the resultant network has approximately the same number of parameters as our proposed network (Fig. \ref{fig:arch-ablation}). The self-attention mechanism for fusing audio and visual features is crucial to attaining good performance; both the success rate and the average reward drop by nearly $50\%$ when replacing the transformer with an MLP on the scooping task. Despite this drop in performance, the alternative MLP architecture with direct concatenation of features performs comparably with Scratch, the next best baseline on the scooping task which shares the same architecture as our method. Hence, the attention mechanism is a necessary yet insufficient condition for attaining good performance when using both visual and contact audio observations---the attention mechanism combined with pre-trained audio and visual features results in favorable performance in the low-data regime.

%% file: source/conclusion.tex
We present a simple yet effective approach for improving manipulation performance by utilizing contact microphones as a tactile sensor. We argue that a primary strength of this sensor is that, in contrast to other sensors, it allows us to leverage large-scale internet data of the same modality and pretrain a representation that is useful for downstream robotic tasks. We show that the representations learned from large-scale audio-visual pretraining transfer well to such tasks despite the domain gap between contact audio in robotic manipulation and audio in internet videos. Future work may investigate which properties of pre-training datasets are most conducive to learning audio-visual representations for manipulation policies. Further, a promising direction would be to equip end-effectors with visuotactile sensors and contact microphones with pre-trained audio representations to determine how to leverage both for equipping robotic agents with a richer understanding of their environment.

The lessons learned from our experiments echo those being shared across other machine learning subfields---more data is the driving factor in learning better models. Considering the safety issues, inefficiency, and resources required in collecting robotic data, it is unlikely that robotics will experience the scaling properties witnessed in more data-rich domains \cite{brown2020language, wei2022emergent}. Thus, we hope to widen the data scarcity bottleneck via methods that extract information from broader data sources that may be useful to an embodied agent.

%% file: source/limitations.tex
While contact microphones work well in our experiments, there are cases in which they may be less useful: less dynamic tasks such as pick and place, situations where the robot itself generates significant vibrations or cases where the robot is working with deformable objects that do not emit perceptible vibrations upon contact.

%% file: source/appendix.tex
\subsection{Architecture}
 The policy takes as input $4$ images and a single audio clip spanning a two-second window, resulting in $5$ total tokens passed to the transformer. Learned positional encodings are applied to the audio and visual features. We use a single self-attention block that follows the traditional transformer encoder structure of \cite{vaswani2017attention}, except that we use pre-layernorm instead of post-layernorm. The self-attention block uses an embedding dimension of $512$, $8$ attention heads, and an expansion ratio of $1$. The output of the transformer block is concatenated and passed to a 2-layer MLP with hidden dimensions of 512. We use a Dropout probability of $0.5$ for all linear layers. The resultant network has around $20$M parameters. Setting $h=2$ at inference strikes a balance between handling the non-Markovian nature of demonstrations and remaining reactive to changes in audio input.

\subsection{Training}
All behavior cloning policies are trained with a batch size of 64 for a maximum of 100 epochs using early stopping with a patience of 15 epochs. We choose the model with the lowest validation loss for evaluation. Pre-trained parameters remain unfrozen during policy learning to mitigate the distribution shift between pre-training data and in-domain data. However, we perform an ablation in Section \ref{sec:zero-shot-transfer} demonstrating that keeping the AVID audio encoder frozen yields only slightly worse results and still outperforms the next best baseline for the zipping task. We apply image augmentations during training with probability $0.5$. When image augmentations are applied, we use PyTorch RandomCrop to size 224 and ColorJitter with the following parameters: brightness $0.3$, contrast $0.3$, saturation $0.1$, and hue $0.2$. 

We use an Adam \cite{kingma2014adam} optimizer and a cosine annealing learning rate scheduler with a starting learning rate of $0.001$. We train on a single GPU (3080Ti) which takes $1$-$1.5$ hours per model. For the BYOL-A baseline, we train a model for each task on the corresponding audio spectrograms for $100$ epochs with a batch size of $1024$, a learning rate of $0.0003$, and the default settings for the network parameters and augmentations from \cite{niizumi2021byol}.